# Neighborhood Rank Order Coding for Robust Texture Analysis and Feature Extraction

Christian Mayr and Rene Schüffny
*University of Technology Dresden, Chair for Neural Circuits and Parallel VLSI-Systems*
*{mayr,schueffn}@iee.et.tu-dresden.de*

**Abstract**

*Research into the visual cortex and general neural information processing has led to various attempts to integrate pulse computation schemes in image analysis systems. Of interest is especially the robustness of representing an analogue signal in the phase or duration of a pulsed, quasi-digital signal, as well as the possibility of direct digital interaction, i.e. computation, among these signals. Such a computation can also achieve information compaction for subsequent processing stages. By using a pulse order encoding scheme motivated by dendritic pulse interaction, we will show that a powerful low-level feature and texture extraction operator, called Pulsed Local Orientation Coding (PLOC), can be implemented. Feature extraction results are being presented, and a possible VLSI implementation is detailed.*

## 1. Introduction

In today's integrated vision systems, their speed, accuracy, power consumption, and complexity depend primarily on the first stage of visual information processing. The task for the first stage is to extract relevant features from an image such as textures, lines and their angles, edges, corners, intersections, etc. These features have to be extracted robustly with respect to illumination, scale, relative contrast, etc. The feature extraction has to be fast, and low in power consumption which can best be achieved by a parallel, mixed-signal implementation [1]. However, the downside of coding the feature extraction in hardware are severe limitations as to flexibility of the features with regard to different applications [1,2], whereas software-based feature extractions could simply be partially reprogrammed to suit different applications. A possible compromise would be a sensor which extracts a very general yet high-quality set of features, with the higher level processing done in software or on a second IC [1,3].

Taking inspiration from the dendritic interaction in biological neural nets [4], where pulses can block or facilitate their counterparts from neighboring neurons, we propose a scheme where free running photo-current driven oscillators [5] interact via their respective pulse trains with their pixel neighbors to form localized topology maps. We will show that these maps contain significant local image content, while at the same time reducing redundancy. This scheme is also somewhat similar to retinal processing in its distributed, localized nature and processing across a small hierarchy of layers. First, the basic tenets of the PLOC operator are detailed, followed by a description of possible post-processing schemes and several examples. The effect of jitter of the pixel pulse frequencies on the PLOC features is discussed and a possible VLSI implementation is described.

## 2. PLOC operator

### 2.1. Basic concept

Employing relative pulse order for information analysis has been postulated in [6] for biological neural nets. How a conventional localized image operator can be implemented using pulse order computation was introduced in [5]. Basically, a hypothesis about the local structure of an image stimulus is generated from single pulsing pixel cells[1] and their relative pulse statistics. In the PLOC operator, these statistics are reduced to the following: A pixel cell resets its registers with a new pulse from itself and then stores pulses from neighbors which occur before its own next pulse (In the following, the interval between two spikes of the same cell will be called Interspike Interval (ISI), per neurobiological convention). Once a new pulse from the center cell occurs, it again resets its registers and transmits its register states using the coefficients given in Fig. 1:

$$\begin{pmatrix} 0 & 1 & 0 \\ 2 & R & 4 \\ 0 & 8 & 0 \end{pmatrix} \begin{pmatrix} 1 & 2 & 4 \\ 8 & R & 16 \\ 32 & 64 & 128 \end{pmatrix} \quad \begin{array}{c} n \\ \uparrow \\ \rightarrow m \end{array}$$

---

[1] For the concept discussion, an arbitrary center pixel will be defined, whose neighboring pixel pulses are analyzed. This constitutes a thought model and does not single out a certain pixel cell, the PLOC cell has identical structures across the pixel matrix.

**Fig. 1: Orientation coefficients of PLOC for N4 and N8 neighborhood**

Thus, if the PLOC operator uses a N4 neighborhood, i.e. considers only pulses from its immediate neighbors, a feature number of 3 would be transmitted if the top and left neighbors pulse within one ISI of the middle PLOC cell. The PLOC cell produces the above estimate of the local image structure for every one of its own ISIs. The operation carried out by the PLOC operator can be derived from the relative phase durations. Starting from an arbitrary time point at the center cell, the probability density function (pdf) for one spike of a neighboring pixel with a regular ISI of duration $T_2$ and arbitrary initial phase to occur can be written as:

$$f_{T_2}(t) = \begin{cases} \dfrac{1}{T_2} & \text{for} \quad 0 \le t < T_2 \\ 0 & \text{else} \end{cases} \quad (1)$$

By integrating this pdf for the ISI duration $T_1$ of the center cell, the probability of a spike of the neighboring cell occurring within one ISI of the center cell can be computed:

$$P(0 \le t \le T_1) = \int_0^{T_1} f_{T_2}(t)dt = \begin{cases} \dfrac{T_1}{T_2} \text{ resp. } \dfrac{\lambda_2}{\lambda_1} & \text{for } T_2 > T_1 \\ 1 & \text{for } T_2 \le T_1 \end{cases} \quad (2)$$

The probability of the corresponding pulse register bit being set hence depends on the ratio of the respective pulse rates. All individual pairings between center cell and a neighbor will produce a set feature bit with the above ratio. Since the initial phase of all cell pulse trains is assumed as random, the phase orders of the neighbors interchange with respect to one another. All possible permutations of neighbor pulses within one ISI of the middle pixel cell (i.e. PLOC features) can thus occur. The probabilities for the respective PLOC features incorporating all N4 or N8 neighbors can accordingly be derived by superposition of individual pulse pairings, i.e. by multiplying the individual probabilities. A sample PLOC feature output distribution for an arbitrary center pulse rate $\lambda_0$ and a sample of grayscale-equivalent pulse rates normalized to this center frequency could accordingly look like Fig. 2:

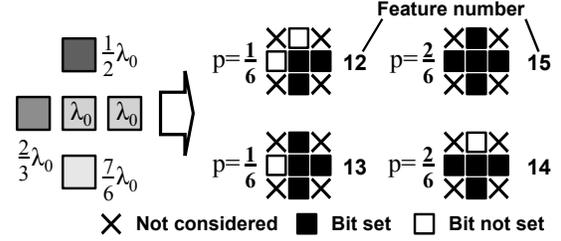

**Fig. 2: PLOC features resulting from a sample pixel pulse frequency distribution**

The absolute pulse rates are not relevant for the probabilities of individual PLOC features, since they cancel out due to the division in equation 2. To extend the above considerations to Fig. 2, we can consider the probability of feature 13. The pixel below the middle pixel has a higher pulse rate than the middle pixel and will thus always produce at least one spike per ISI of the middle pixel. The 8-valued bit of the N4 neighborhood will thus always be set. The same is true for the pixel to the right of the middle pixel. The pixel above the middle pixel will produce a pulse for half of the ISIs of the middle pixel. In the same way, the pixel to the left will produce no pulse for one third of the ISIs. The probability of feature 13 can be computed as those ISIs where the top pixel pulses while the pixel to the left is silent. If, as mentioned above, the initial phases of the pixels are assumed to be uncorrelated, this would be true for 1/6 of the total ISIs.

As can be seen, dominant and secondary local image structure is captured by the PLOC features. Compared with a more conventional operator using some kind of local thresholding, the PLOC operator achieves a kind of interrelated judgment of image structure. For example, a threshold operator could not signify feature 13. Because the pulsing pixel cells are assumed to produce regular rates with constant ISIs, the above probability notation is somewhat misleading. Even for a small sample of ISIs of the center cell, the features cycle deterministically through the above permutations.

Therefore, all feature information is captured within a small time span, making for rapid image analysis. Fig. 3 gives an impression what kind of information is captured by PLOC (All pixels which have transmitted feature 7 at least once are marked in black). The PLOC simulation results detailed below and in the rest of this paper assume a linear conversion between pixel grayscale value and pulse frequency $\lambda$. This can be achieved easily with simple CMOS photo diodes, thresholding and reset [5,7].

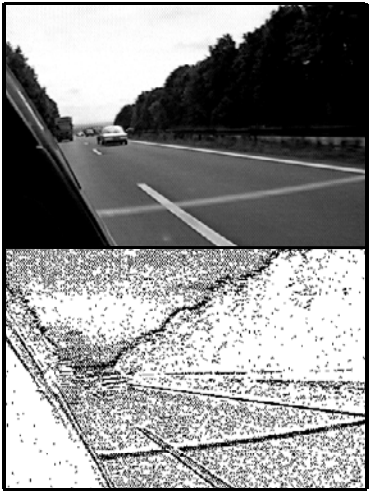

**Fig. 3: Original image and basic PLOC feature number 7**

Despite operating on a very small neighborhood, significant large-scale image structure is captured by the operator. In the above form, however, the features are not selective enough and could profit from postprocessing.

## 2.2. Postprocessing of PLOC features

The first kind of postprocessing is a significance assessment of the individual PLOC features. For every PLOC cell at coordinate (m,n), a time frame of length T delivers λ*T features in total. A normalization of the occurrence $N_k$ of each individual feature k by this total number delivers a measure for its probability as in Fig. 2. Subjecting this to an appropriate threshold $\theta_M$ will weed out features of a more stochastic, intermittent nature in favor of those containing more systematic information about the local image structure:

$$b_k^{'}(m,n) = \begin{cases} 1 & for \quad \dfrac{N_k(m,n)}{\sum_i N_i(m,n)} \geq \theta_M \\ 0 & else \end{cases} \quad (3)$$

This way, the features returned by the PLOC operator could be restricted to 14 and 15 for the example given in Fig. 2 for a $\theta_M$ of 0.2. As can be seen from Fig. 2, the changing relative phase orders cause the image to be analyzed along several different flavors of local image information. Accordingly, several PLOC features can contain significant image information, so the above threshold assessment is carried out for all PLOC features at every coordinate, resulting in a feature vector containing an entry of one for all the dominant features at the location (m,n):

$$b'(m,n) = \begin{pmatrix} b_1^{'}(m,n) \\ b_2^{'}(m,n) \\ \vdots \\ b_k^{'}(m,n) \end{pmatrix} \quad (4)$$

In other words, a batch of images is created, with each one containing the answer for a single feature across the original image.

An additional improvement in the PLOC feature quality can be achieved through a spatial correlation analysis. PLOC features which are indicative of large scale image structures should be discernible through a localized accumulation (see lower half of Fig. 3). This localized clustering could be analyzed by comparing the feature vectors b'(m,n) of neighboring PLOC cells, which should be partially correlated. A correlation measure is defined which computes the similarity between the center pixel and one neighbor:

$$A(m+i,n+j) = \begin{cases} 1 & for \quad \dfrac{\sum_k [b_k^{'}(m,n) \cap b_k^{'}(m+i,n+j)]}{\sum_k [b_k^{'}(m,n) \cup b_k^{'}(m+i,n+j)]} \geq \theta_{corr} \\ 0 & else \end{cases} \quad (5)$$

The intersection among elements from the feature vector is divided by the respective set union and evaluated by a similarity threshold $\theta_{corr}$. The relative coordinates (i,j) are permutated across the complete N8 neighborhood of the operator. The set k of single feature numbers which are employed in computing the correlation measure is not necessarily the complete PLOC feature set, rather an adjusted one dependent on the tasks at hand. For example, to extract the upper border of the street middle line in Fig. 3, k would contain features such as 11 and 14 of the N4 neighborhood (see also Fig. 4). From the pair wise correlations A(m+i,n+j) between the center pixel and one neighbor b(m+i,n+j), an overall assessment of the correlation between a pixel and its neighbors is computed:

$$b_{korr}(m,n) = \begin{cases} 1 & for \quad \sum_{i,j} A(m+i,n+j) \geq N_{corr} \\ 0 & else \end{cases} \quad (6)$$

Via its threshold $N_{corr}$, this thresholded sum of A(m+i,n+j) captures pixels belonging to a single large-scale feature, because their respective feature vectors should be heavily correlated.

A few examples for the two types of postprocessing are shown in Fig. 4. If a normalization according to equation 3 is used, with a threshold $\theta_M$ of 0.1, the image in the upper left can be obtained, with a positive threshold decision once again denoted in black. A significant improvement can be observed in the reduction of spurious PLOC features for uniform areas such as the tarmac. The sky seems to have significant small scale structure and does not respond as well to the significance assessment. Using the correlation operator with just feature 7 and $N_{corr}$ of 5, the upper right image shows reduced features for the sky. In this usage, the correlation operator is reduced to a decision whether a majority of surrounding pixels exhibits an identical feature.

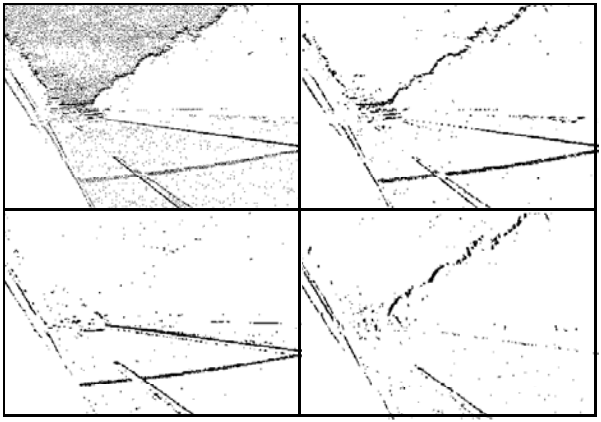

**Fig. 4: Postprocessing with significance thresholding of $\theta_M$=0.1 (top left, feature 7) and additional neighborhood correlation (clock-wise from top right features 7, 11, 14)**

If a single feature is employed instead of a feature set for the correlation operator, the setting of $\theta_{corr}$ is not critical. Since the correlation is reduced to a comparison, the operation is carried out correctly for any $\theta_{corr}$ greater zero.

Of interest is the robustness with respect to absolute grayscale levels and border contrast which the PLOC features exhibit in the above figure. Some of the macroscopic features show themselves for several PLOC features, e.g. the upper border of the forest/greenery in the upper two images in Fig. 4. Other large-scale features, like the middle line of the road, show up for several PLOC features, but are analyzed slightly different for each one, with right border, let border, or both being exhibited. High-level symbolic postprocessing could use this information delivered by the PLOC operator to e.g. extract edge contrast direction. Also, interpolation of edge orientation could be carried out among the fundamental angles of the single PLOC features where the edge shows up.

### 2.3. Additional simulation results

A combination of certain PLOC features can also be used for extracting a general set of high-interest points from an image. For the following demonstration, features of the N4 neighborhood such as lines (6, 9), end points of lines (1, 2, 4, 8), and corners (3, 5, 10, 12) are used as feature subvector. With a setting of $\theta_M$=0.1 and $\theta_{corr}$=0.3, these significant single points can be extracted from a test image:

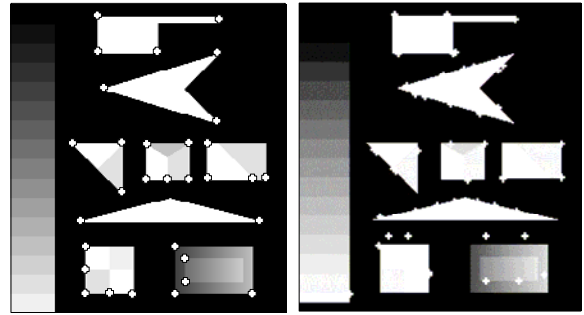

**Fig. 5: Salient points extracted by PLOC (left) compared with a wavelet based method [3]**

The salient points extracted by PLOC are comparable with the ones derived by using a much more computationally expensive wavelet operator [3]. The parameter settings as above are actually not critical for this artificial test image. However, this salient points analysis was also carried out with natural images, the parameters represent the settings which work best for these more realistic input images.

## 3. Pixel jitter repercussions

For an implementation, but also with respect to neural information processing, it would be of interest to evaluate how much noise/jitter degrades the PLOC feature response. For example, a pulse rate jitter could cause sporadic pulse time shifts despite a stable overall frequency and could thus cause feature artifacts. In the following, an exemplary analysis is carried out for identical pulse rates of middle pixel and one neighbor. According to equation 2, this should cause a 'definite' set bit/partial feature, since there is a pulse of the neighbor pixel for every pulse of the middle pixel. However, jitter could cause two of the neighbor pulses to fall within one pulse period of the middle pixel and accordingly miss the next period of the middle pixel. For this analysis, jitter of the middle pixel and the

neighbor pixel are combined in the jitter of the relative phase difference between both pixels:

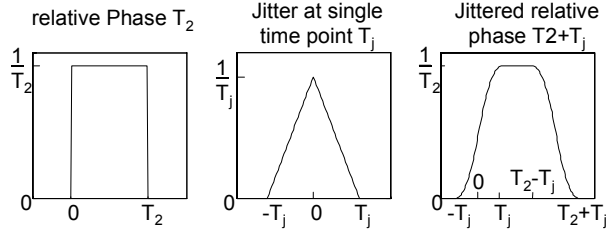

**Fig. 6: Relative phase, (assumed) jitter, and resultant jittered relative phase of a pixel pulse**

For simplicity, it is assumed that the pulses from pixel 2 are jittered with respect to their true time of occurrence according to the following triangular pdf:

$$f_{T_j}(t) = \begin{cases} \dfrac{(t+T_j)}{T_j^2} & \text{for } -T_j \leq t < 0 \\ -\dfrac{(t-T_j)}{T_j^2} & \text{for } 0 \leq t < T_j \\ 0 & \text{else} \end{cases} \quad (7)$$

The pdf of the jittered version of pixel 2 can be derived from the addition of the variates of equations (1) and (7), i.e. from the convolution of their pdfs. To satisfy all constraints from both equations, the convolution integral has to be subdivided in five sections and parameter-dependent integral-borders have to be used. Since this analysis is only concerned with pulses jittered outside interval $T_2$, only the first of the five sections will be treated here:

$$f_{T_2 T_j}(t) = \int_0^{T_j+t} f_{T_2}(\tau) * f_{T_j}(t-\tau) d\tau \text{ for } -T_j \leq t < 0 \quad (8)$$

Carrying out this integration gives the following pdf for this section:

$$f_{T_2 T_j}(t) = \frac{1}{T_2 T_j^2}\left[-\frac{\tau^2}{2} + (t+T_j)\tau\right]_0^{T_j+t} = \frac{(t+T_j)^2}{T_2 T_j^2} \quad (9)$$

From the above equation, the probability of a bit being erroneously omitted can be derived as:

$$P(t \leq 0 \cup t \geq T_2) = 2\int_0^{T_j} \frac{(t+T_j)}{2T_2 T_j^2} dt = \\ = \frac{1}{3T_2 T_j^2}\left[(t+T_j)^3\right]_{-T_j}^0 = \frac{T_j}{3T_2} \quad (10)$$

Since the pdf is symmetric, a factor of two is incorporated in the above equation to take into account both ends of the interval $T_2$. A (biological) pulse rate of $50s^{-1}$ i.e. a $T_2$ of 20 ms with a jitter of 1 ms [8] would result in an error probability of 1,7%. For technical implementations of such schemes, the ratio between jitter und pulse rate is of the same order of magnitude or better (e.g. 10kHz pixel pulse frequency and 2-4μs jitter of the integrator). Errors among PLOC features caused by jitter are therefore well below any reasonable significance assessment and can be eliminated through an appropriate choice of $\theta_M$.

## 4. VLSI implementation

A VLSI implementation of the PLOC cell using the pulsing pixel cell as detailed in [5] could be easily carried out using a small number of standard digital gates with the following structure for an N4 neighborhood:

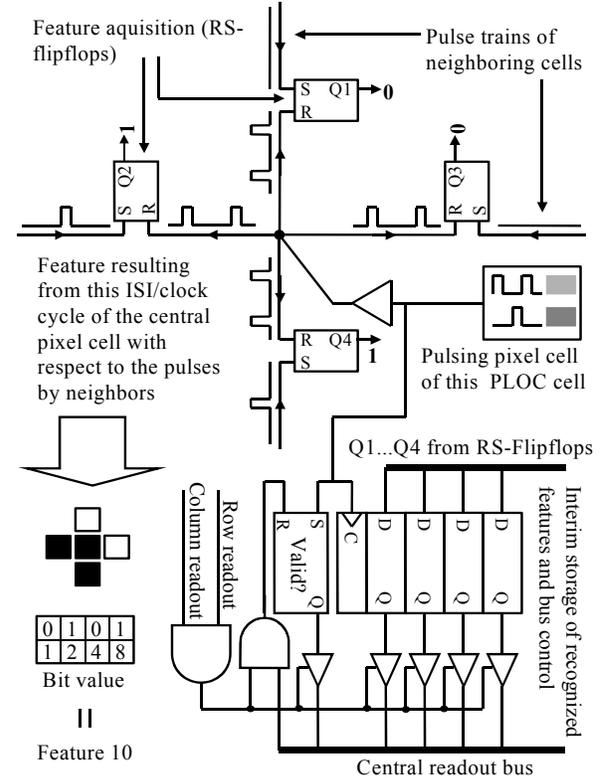

**Fig. 7: PLOC features resulting from a sample pixel pulse frequency distribution**

The first step would be a number of RS-latches acquiring the neighboring pulses, with a set of slave latches in series to store results from the last ISI for readout. Transmission of memory contents to the slaves and reset of the RS-latches is carried out for every pulse of the center pixel cell, with the reset slightly delayed to ensure secure transmission to the slaves.

Bus drivers, column and row select complete the implementation. Since the features are generated asynchronously at the pixel pulse rate, a simple choice for readout would be by scanning all PLOC cells at the maximum pixel pulse rate via a central bus. In this case, an additional valid bit is necessary to avoid duplicate readout of a feature. The feature normalization as per equation (3) and (4) can be realized in feature accumulators at the edge of the PLOC matrix. It would be carried out once a $2^N$ number of overall features for every PLOC cell is stored in the accumulators, thus limiting the normalization of individual feature numbers to an N bit shift operation. Those normalized individual feature numbers are then subjected to a global threshold $\theta_M$ and readout from the IC.

For a complete N8 neighborhood, the area requirements of the above implementation in a 130 nm CMOS technology can be estimated as follows: Standard digital building blocks 8 D-FF, 9 RS-FF, 2 AND, 9 tristate drivers, and one buffer. With an estimated digital fill factor of 80%, this results in an chip area of 872 $\mu m^2$. Additional 172 $\mu m^2$ are needed for the pulsing pixel cell, giving a square PLOC cell with 32 $\mu m$ border length. A comparable IC realization of a somewhat similar conventional image analysis operator, called Local Orientation Coding (LOC) [5] in a 0.6 $\mu m$ process resulted in a cell size of 83$\mu m$*80$\mu m$. Minimum structure size of the technology is not an issue in this comparison, since most components of the LOC cell were analog and thus do not shrink with the technology.

## 5. Conclusion

As can be seen from Fig. 4, the postprocessed PLOC features can be effectively used as basis for a more symbolic, high level image analysis. At the same time, they avoid the usual pitfall of hardware coded image analysis, which is often too specialized, loosing image information the preprocessing sensor is not geared for. Additionally, large image areas without significant image information are blanked, reducing the processing load on subsequent processing stages.

The tendency of the PLOC operator to deliver the predominant local features even in the presence of an arbitrary initial phase may also be of neurobiological relevance: VanRullen et al. analyze a Rank Order Coding (ROC) in [9] that uses 'Time to first Spike' as the relevant information carrying attribute of pulse trains. However, they give no hint about any biological effect which could be the necessary reset mechanism to ensure a faithful decoding of the first spike after stimulus change. Since the system treated in the paper (Retina up to first stage of V1 in the visual cortex) does not possess such a feedback path apart from maybe saccades, the reset in [9] is questionable. As has been indicated in the previous sections, the online, continuous ROC employed for the PLOC operator is able to capture the dominant phase relationship within very few ISIs of the corresponding pulse trains without reset. This type of quasi-dendritic interaction might thus be also at work in neural tissue, alleviating the need for a reset or feedback path.

## 6. Acknowledgment

The authors would like to thank Prof. Andreas König for initially pointing out the LOC operator and for many fruitful discussions during its refinement. Also, Prof. König provided valuable assistance in drafting this manuscript.